\pdfoutput=1
\documentclass[10pt,twocolumn,letterpaper]{article}

\usepackage{cvpr}
\usepackage{times}
\usepackage{epsfig}
\usepackage[utf8]{inputenc} % allow utf-8 input
\usepackage[T1]{fontenc}    % use 8-bit T1 fonts
\usepackage{url}            % simple URL typesetting
\usepackage{booktabs}       % professional-quality tables
\usepackage{amsfonts}       % blackboard math symbols
\usepackage{amsmath}       
\usepackage{amssymb}       
\usepackage{nicefrac}       % compact symbols for 1/2, etc.
\usepackage{microtype}      % microtypography
\usepackage{color}
\usepackage{graphicx}
\usepackage{floatrow}
\usepackage{tabularx}
\usepackage{hyperref}       % hyperlinks

\cvprfinalcopy

\newfloatcommand{capbtabbox}{table}[][\FBwidth]

\ifcvprfinal\fi
\begin{document}

\title{Associative Embedding: \\ End-to-End Learning for Joint 
  Detection and Grouping}

\author{
  Alejandro Newell \\ University of Michigan \\ \tt\small alnewell@umich.edu
  \and
  Zhiao Huang* \\ Tsinghua University \\ \tt\small hza14@mails.tsinghua.edu.cn
  \and
  Jia Deng \\ University of Michigan \\ \tt\small jiadeng@umich.edu
}

\maketitle

%%%%%%%%% ABSTRACT
\begin{abstract}
  We introduce associative embedding, a novel method for supervising
  convolutional neural networks for the task of detection and
  grouping. A number of computer vision problems can be framed in this
  manner including multi-person pose estimation, instance
  segmentation, and multi-object tracking. Usually the grouping of
  detections is achieved with multi-stage pipelines, instead we
  propose an approach that teaches a network to simultaneously output
  detections and group assignments. This technique can be easily
  integrated into any state-of-the-art network architecture that
  produces pixel-wise predictions. We show how to apply this method to
  both multi-person pose estimation and instance segmentation and
  report state-of-the-art performance for multi-person pose on the
  MPII and MS-COCO datasets.

\end{abstract}

\renewcommand*{\thefootnote}{\fnsymbol{footnote}}
\footnotetext{* Work done while a visiting student at the University of Michigan.}

%%%%%%%%% BODY TEXT
\section{Introduction}

Many computer vision tasks can be viewed as joint detection and
grouping: detecting smaller visual units and grouping them into larger
structures. For example, multi-person pose estimation can be viewed as
detecting body joints and grouping them into individual people; instance
segmentation can be viewed as detecting relevant pixels and grouping
them into object instances; multi-object tracking can be viewed as
detecting object instances and grouping them into tracks. In all of these
cases, the output is a variable number of visual units and their
assignment into a variable number of visual groups.

\begin{figure}[t]
\centering
\includegraphics[width=\textwidth, trim={0 .5cm 0 2.5cm}, clip]{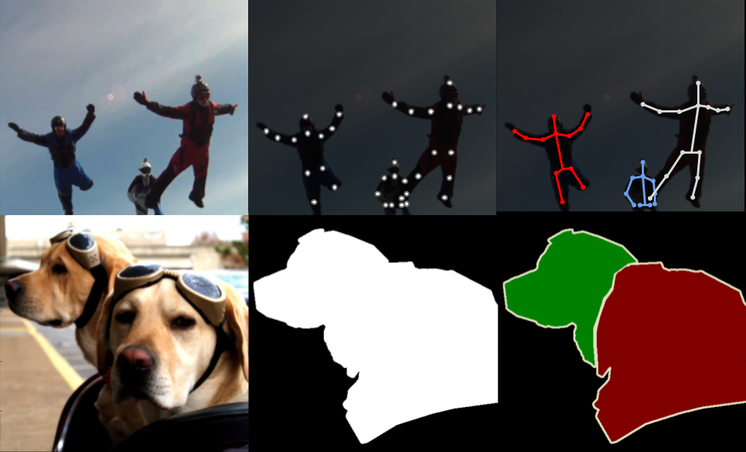}
\caption{Both multi-person pose estimation and instance segmentation are examples of computer vision tasks that require detection of visual elements (joints of the body or pixels belonging to a semantic class) and grouping of these elements (as poses or individual object instances).}
\label{fig:intro-fig}
%\vspace{-.2cm}
\end{figure}

Such tasks are often approached with two-stage pipelines that
perform detection first and grouping second. But such approaches may
be suboptimal because detection and grouping are usually tightly
coupled: for example, in multiperson pose estimation, a wrist
detection is likely a false positive if there is not an elbow
detection nearby to group with.

In this paper we ask whether it is possible to jointly perform
detection and grouping using a single-stage deep network trained
end-to-end. We propose \textit{associative embedding}, a novel method
to represent the output of joint detection and grouping. The basic
idea is to introduce, for each detection, a real number that serves as
a ``tag'' to identify the group the detection belongs to. In other
words, the tags associate each detection with other detections in the same group.

Consider the special case of detections in 2D and embeddings in 1D
(real numbers). The network outputs both a heatmap of per-pixel
detection scores and a heatmap of per-pixel identity tags. The
detections and groups are then decoded from these two heatmaps.

To train a network to predict the tags, we use a loss function that
encourages pairs of tags to have similar values if the corresponding
detections belong to the same group in the ground truth or dissimilar
values otherwise. It is important to note that we have no “ground
truth” tags for the network to predict, because what matters is not
the particular tag values, only the differences between them.  The
network has the freedom to decide on the tag values as long as they
agree with the ground truth grouping.

We apply our approach to multiperson pose estimation, an important
task for understanding humans in images. Concretely, given an input
image, multi-person pose estimation seeks to detect each person and
localize their body joints. Unlike single-person pose there are no
prior assumptions of a person’s location or size. Multi-person pose
systems must scan the whole image detecting all people and their
corresponding keypoints. For this task, we integrate associative
embedding with a stacked hourglass network [31], which produces a
detection heatmap and a tagging heatmap for each body joint, and then
groups body joints with similar tags into individual
people. Experiments demonstrate that our approach outperforms all
recent methods and achieves state of the art results on MS-COCO [27]
and MPII Multiperson Pose [3, 35].

We further demonstrate the utility of our method by applying it to
instance segmentation. Showing that it is straightforward to apply
associative embedding to a variety of vision tasks that fit under the
umbrella of detection and grouping.

Our contributions are two fold:
(1) we introduce associative embedding, a new method for single-
stage, end-to-end joint detection and grouping. This method is simple
and generic; it works with any network architecture that produces
pixel-wise prediction; (2) we apply associative embedding to
multiperson pose estimation and achieve state of the art results on
two standard benchmarks.

\section{Related Work}

\noindent
{\bf Vector Embeddings} Our method is related to many prior works that
use vector embeddings. Works in image retrieval have used vector
embeddings to measure similarity between
images~\cite{frome2007learning,weinberger2005distance}. Works in image
classification, image captioning, and phrase localization have used
vector embeddings to connect visual features and text features by
mapping them to the same vector
space~\cite{frome2013devise,gong2014improving,karpathy2015deep}. Works
in natural language processing have used vector embeddings to
represent the meaning of words, sentences, and
paragraphs~\cite{mikolov2013distributed,le2014distributed}. Our work
differs from these prior works in that we use vector embeddings as
identity tags in the context of joint detection and grouping.

\smallskip
\noindent
{\bf Perceptual Organization} Work in perceptual organization aims to
group the pixels of an image into regions, parts, and
objects. Perceptual organization encompasses a wide range of tasks of
varying complexity from figure-ground
segmentation~\cite{maire2010simultaneous} to hierarchical image
parsing~\cite{han2009bottom}. Prior works typically use a two stage
pipeline~\cite{maire2011object}, detecting basic visual units
(patches, superpixels, parts, etc.) first and grouping them
second. Common grouping approaches include spectral
clustering~\cite{von2007tutorial,shi2000normalized}, conditional
random fields (e.g.\@~\cite{koltun2011efficient}), and generative
probabilistic models (e.g.\@~\cite{han2009bottom}). These grouping
approaches all assume pre-detected basic visual units and pre-computed
affinity measures between them but differ among themselves in the
process of converting affinity measures into groups. In contrast, our
approach performs detection and grouping in one stage using a generic
network that includes no special design for grouping.

It is worth noting a close connection between our approach to those
using spectral clustering. Spectral clustering (e.g.\@ normalized
cuts~\cite{shi2000normalized}) techniques takes as input pre-computed
affinities (such as predicted by a deep network) between visual units
and solves a generalized eigenproblem to produce embeddings (one per
visual unit) that are similar for visual units with high
affinity. Angular
Embedding~\cite{maire2010simultaneous,stella2009angular} extends
spectral clustering by embedding depth ordering as well as grouping.
Our approach differs from spectral clustering in that we have no
intermediate representation of affinities nor do we solve any
eigenproblems. Instead our network directly outputs the final
embeddings. 

Our approach is also related to the work by Harley et al. on learning dense convolutional
embeddings~\cite{harley2015iclr}, which trains a deep network to produce pixel-wise
embeddings for the task of semantic segmentation. Our work differs from theirs in that our
network produces not only pixel-wise embeddings but also pixel-wise detection scores. Our
novelty lies in the integration of detection and grouping into a single network; to the
best of our knowledge such an
integration has not been attempted for multiperson human pose estimation.

\begin{figure*}[t]
\centering
\includegraphics[width=.85\textwidth, clip]{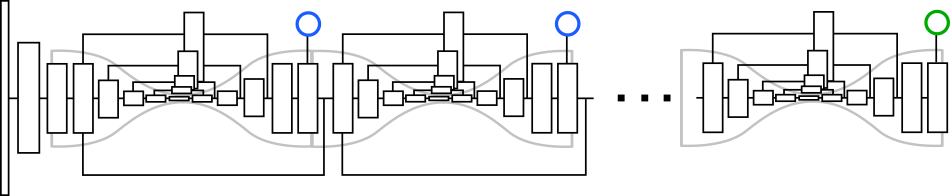}
\caption{We use the stacked hourglass architecture from Newell et
  al. \cite{newell2016stacked}. The network performs repeated
  bottom-up, top-down inference producing a series of intermediate
  predictions (marked in blue) until the last ``hourglass'' produces a
  final result (marked in green). Each box represents a 3x3
  convolutional layer. Features are combined across scales by
  upsampling and performing elementwise addition. The same ground
  truth is enforced across all predictions made by the network. }
\label{fig:stacked-hg}
\end{figure*}

\smallskip
\noindent
{\bf Multiperson Pose Estimation} Recent methods have made great
progress improving human pose estimation in images in particular for
single person pose
estimation~\cite{toshev2014deeppose,tompson2015efficient,wei2016machines,newell2016stacked,
  chu2017multi, bulat2016human, ning2017knowledge,
  belagiannis2016recurrent, fan2015combining, chain16,
  lifshitz2016human, hu2016bottom,carreira2016human, tompson2014joint,
  pishchulin2013poselet}.  For multiperson pose, prior and concurrent
work can be categorized as either top-down or bottom-up.  Top-down
approaches~\cite{papandreou2017towards, he2017mask, fang2016rmpe}
first detect individual people and then estimate each person's
pose. Bottom-up approaches
~\cite{pishchulin16cvpr,insafutdinov16ariv,iqbal2016multi,
  cao2016realtime} instead detect individual body joints and then
group them into individuals. Our approach more closely resembles
bottom-up approaches but differs in that there is no separation of a
detection and grouping stage. The entire prediction is done at once by
a single-stage, generic network. This does away with the need for
complicated post-processing steps required by other methods
\cite{cao2016realtime,insafutdinov16ariv}.

\smallskip
\noindent
{\bf Instance Segmentation} Most existing instance segmentation
approaches employ a multi-stage pipeline to do detection followed
by segmentation
\cite{hariharan2015hypercolumns,girshick2015deformable,hariharan2014simultaneous,dai2015convolutional}. Dai
et al.~\cite{dai2015instance} made such a pipeline differentiable
through a special layer that allows backpropagation through spatial
coordinates.

Two recent works have sought tighter integration of detection and
segmentation using fully convolutional networks.
DeepMask~\cite{pinheiro2015learning} densely scans subwindows and
outputs a detection score and a segmentation mask (reshaped to a
vector) for each subwindow. Instance-Sensitive
FCN~\cite{dai2016instance} treats each object as composed of a set of
object parts in a regular grid, and outputs a per-piexl heatmap of
detection scores for each object part. Instance-Sensitive FCN (IS-FCN)
then detects object instances where the part detection scores are
spaitally coherent, and assembles object masks from the heatmaps of
object parts. Compared to DeepMask and IS-FCN, our approach is
substantially simpler: for each object category we output only two
values at each pixel location, a score representing foreground versus
background, and a tag representing the identity of an object instance,
whereas both DeepMask and IS-FCN produce much higher dimensional
output.

\begin{figure*}[t]
\centering
\includegraphics[width=.9\textwidth, clip]{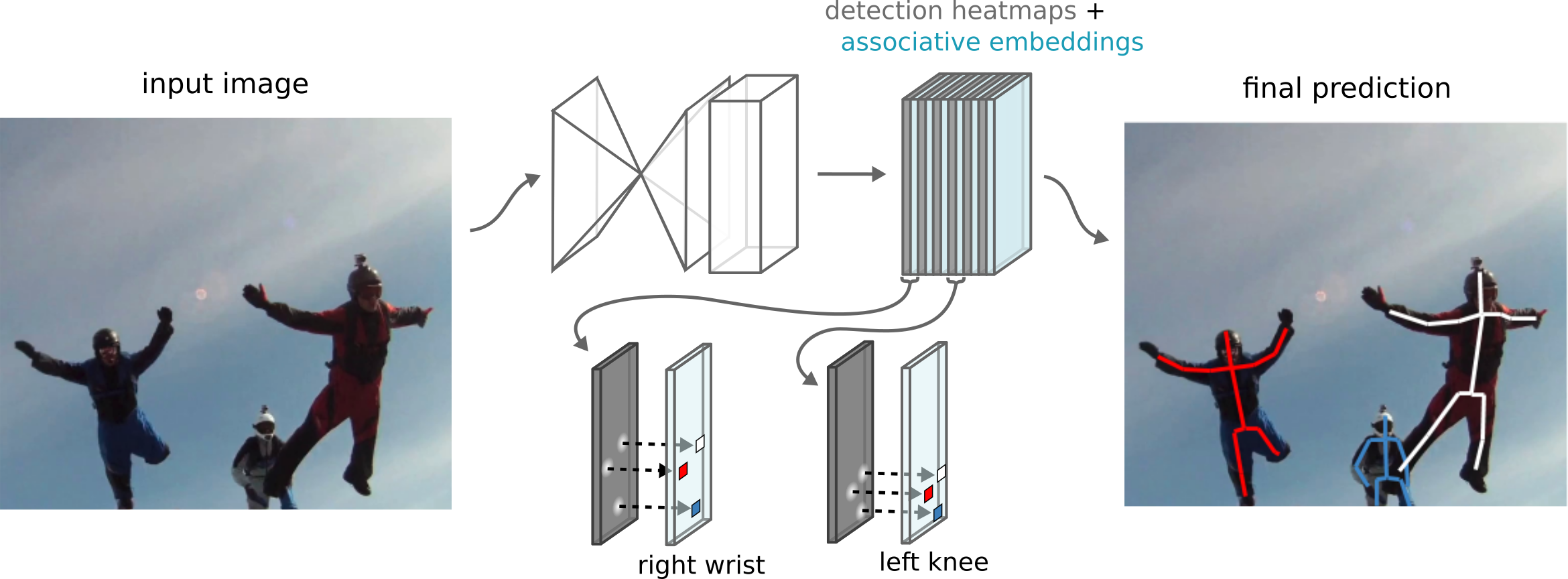}
\caption{An overview of our approach for producing multi-person pose
  estimates. For each joint of the body, the network simultaneously
  produces detection heatmaps and predicts associative embedding
  tags. We take the top detections for each joint and match them to
  other detections that share the same embedding tag to produce a
  final set of individual pose predictions.}
\label{fig:multi-1}
\vspace{-.2cm}
\end{figure*}

\section{Approach}

\subsection{Overview} 
To introduce associative embedding for joint detection and grouping,
we first review the basic formulation of visual detection. Many visual
tasks involve detection of a set of visual units. These tasks are
typically formulated as scoring of a large set of candidates. For
example, single-person human pose estimation can be formulated as
scoring candidate body joint detections at all possible pixel
locations. Object detection can be formulated as scoring candidate
bounding boxes at various pixel locations, scales, and aspect ratios.

The idea of associative embedding is to predict an embedding for each
candidate in addition to the detection score. The embeddings serve as
tags that encode grouping: detections with similar tags should be
grouped together.  In multiperson pose estimation, body joints with
similar tags should be grouped to form a single person. It is
important to note that the absolute values of the tags do not matter,
only the distances between tags.  That is, a network is free to assign
arbitrary values to the tags as long as the values are the same for
detections belonging to the same group.

Note that the dimension of the embeddings is not critical. If a
network can successfully predict high-dimensional embeddings to
separate the detections into groups, it should also be able to learn
to project those high-dimensional embeddings to lower dimensions, as
long as there is enough network capacity. In practice we have found
that 1D embedding is sufficient for multiperson pose estimation, and
higher dimensions do not lead to significant improvement. Thus
throughout this paper we assume 1D embeddings.

To train a network to predict the tags, we enforce a loss that
encourages similar tags for detections from the same group and
different tags for detections across different groups. Specifically,
this tagging loss is enforced on candidate detections that coincide
with the ground truth. We compare pairs of detections and define a
penalty based on the relative values of the tags and whether the
detections should be from the same group. 

\subsection{ Stacked Hourglass Architecture}

In this work we combine associative embedding with the stacked
hourglass architecture~\cite{newell2016stacked}, a model for dense
pixel-wise prediction that consists of a sequence of modules each
shaped like an hourglass (Fig.~\ref{fig:stacked-hg}).  Each
``hourglass'' has a standard set of convolutional and pooling layers
that process features down to a low resolution capturing the full
context of the image. Then, these features are upsampled and gradually
combined with outputs from higher and higher resolutions until
reaching the final output resolution. Stacking multiple
hourglasses enables repeated bottom-up and top-down inference to
produce a more accurate final prediction. We refer the reader to
\cite{newell2016stacked} for more details of the network
architecture.

The stacked hourglass model was originally developed for single-person
human pose estimation. The model outputs a heatmap for each body joint
of a target person. Then, the pixel with the highest heatmap
activation is used as the predicted location for that joint. The
network is designed to consolidate global and local features which
serves to capture information about the full structure of the body
while preserving fine details for precise localization. This balance
between global and local features is just as important in other
pixel-wise prediction tasks, and we therefore apply the same network
towards both multiperson pose estimation and instance segmentation.

We make some slight modifications to the network architecture. We
increase the number of ouput features at each drop in resolution (256 ->
386 -> 512 -> 768). In addition, individual layers are composed of 3x3
convolutions instead of residual modules, the shortcut effect to ease
training is still present from the residual links across each
hourglass as well as the skip connections at each resolution.

\subsection{Multiperson Pose Estimation}

To apply associative embedding to multiperson pose estimation, we
train the network to detect joints as performed in single-person pose
estimation \cite{newell2016stacked}. We use the stacked hourglass
model to predict a detection score at each pixel location for each
body joint (``left wrist'', ``right shoulder'', etc.) regardless of
person identity. The difference from single-person pose being that an
ideal heatmap for multiple people should have multiple peaks (e.g.\@
to identify multiple left wrists belonging to different people), as
opposed to just a single peak for a single target person.

In addition to producing the full set of keypoint detections, the
network automatically groups detections into individual poses. To do
this, the network produces a tag at each pixel location for each
joint. In other words, each joint heatmap has a corresponding ``tag''
heatmap. So, if there are $m$ body joints to predict then the network
will output a total of $2m$ channels, $m$ for detection and $m$ for
grouping. To parse detections into individual people, we use
non-maximum suppression to get the peak detections for each joint and
retrieve their corresponding tags at the same pixel location
(illustrated in Fig. \ref{fig:multi-1}). We then group detections
across body parts by comparing the tag values of detections and
matching up those that are close enough. A group of detections now
forms the pose estimate for a single person.

%training

To train the network, we impose a detection loss and a grouping loss
on the output heatmaps. The detection loss computes mean square
error between each predicted detection heatmap and its ``ground
truth'' heatmap which consists of a 2D gaussian activation at each
keypoint location. This loss is the same as the one used by
Newell et al.~\cite{newell2016stacked}.

The grouping loss assesses how well the predicted tags agree with the
ground truth grouping. Specifically, we retrieve the predicted tags
for all body joints of all people at their ground truth locations; we
then compare the tags within each person and across people. Tags
within a person should be the same, while tags across people should be
different.

Rather than enforce the loss across all possible pairs of keypoints,
we produce a reference embedding for each person. This is done by
taking the mean of the output embeddings of the person's joints. Within an
individual, we compute the squared distance between the reference
embedding and the predicted embedding for each joint. Then, between
pairs of people, we compare their reference embeddings to each other
with a penalty that drops exponentially to zero as the distance
between the two tags increases.

Formally, let $h_k \in R^{W \times H}$ be the predicted tagging
heatmap for the $k$-th body joint, where $h(x)$ is a tag value at
pixel location $x$. Given $N$ people, let the ground truth body joint
locations be $T = \{(x_{nk})\}, n=1,\ldots,N,k=1\ldots,K$, where
$x_{nk}$ is the ground truth pixel location of the $k$-th body joint
of the $n$-th person.

Assuming all K joints are annotated, the reference embedding for the $n$th person would be
$$\bar{h}_n = \frac{1}{K} \sum_k h_k(x_{nk})$$

\begin{figure*}[t]
\centering
\includegraphics[width=\textwidth]{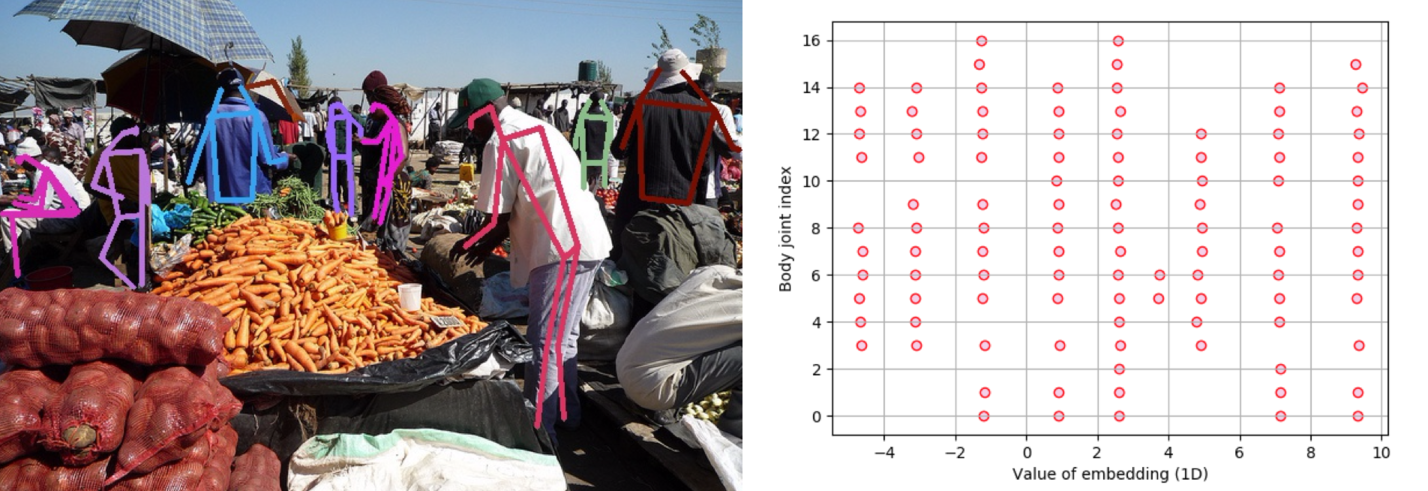}
\caption{Tags produced by our network on a held-out validation image from the MS-COCO training set. The tag values are
  already well separated and decoding the groups is straightforward. }
\label{fig:ae-vs-kpt}
\end{figure*}

The grouping loss $L_g$ is then defined as
\begin{multline*}
 L_g(h,T) =
 \frac{1}{N}\sum_{n}\sum_{k}\left(\bar{h}_{n}
 - h_k(x_{nk},)\right)^2 \\ + \frac{1}{N^2}\sum_{n} \sum_{n'}
 \exp\{-\frac{1}{2\sigma^2}\left(\bar{h}_{n} -
 \bar{h}_{n'} \right)^2\},
\end{multline*}

To produce a final set of predictions we iterate through each joint
one by one. An ordering is determined by first considering joints
around the head and torso and gradually moving out to the limbs. We
start with our first joint and take all activations above a certain
threshold after non-maximum suppression. These form the basis for our
initial pool of detected people.

We then consider the detections of a subsequent joint. We compare the
tags from this joint to the tags of our current pool of people, and
try to determine the best matching between them. Two tags can only be
matched if they fall within a specific threshold. In addition, we want
to prioritize matching of high confidence detections. We thus perform
a maximum matching where the weighting is determined by both the tag
distance and the detection score. If any new detection is not
matched, it is used to start a new person instance. This accounts for
cases where perhaps only a leg or hand is visible for a particular
person.

We loop through each joint of the body until every detection has been
assigned to a person. No steps are taken to ensure anatomical
correctness or reasonable spatial relationships between pairs of
joints. To give an impression of the types of tags produced by the
network and the trivial nature of grouping we refer to Figure \ref{fig:ae-vs-kpt}.

While it is feasible to train a network to make pose predictions for
people of all scales, there are some drawbacks. Extra capacity is
required of the network to learn the necessary scale invariance, and
the precision of predictions for small people will suffer due to
issues of low resolution after pooling. To account for this, 
%we train
%the network on people within a fixed range of sizes, and 
we evaluate images at test time at multiple scales. There are a number of
potential ways to use the output from each scale to produce a final
set of pose predictions. For our purposes, we take the produced
heatmaps and average them together. Then, to combine tags across
scales, we concatenate the set of tags at a pixel location into a
vector $v \in R^m$ (assuming $m$ scales). The decoding process does
not change from the method described with scalar tag values, we now
just compare vector distances.

\begin{figure}
\centering
\includegraphics[width=\textwidth, clip]{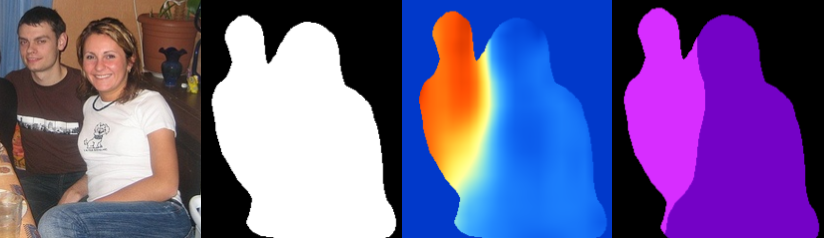}
\caption{To produce instance segmentations we decode the network
  output as follows: First we threshold on the detection heatmap, the
  resulting binary mask is used to get a set of tag values. By looking
  at the distribution of tags we can determine identifier tags for
  each instance and match the tag of each activated pixel to the
  closest identifier.}
\label{fig:inst-hm}
\vspace{-.1cm}
\end{figure}

\begin{figure*}
\centering
\includegraphics[width=.94\textwidth, trim={7.1cm 3.2cm 7.1cm 2cm}, clip]{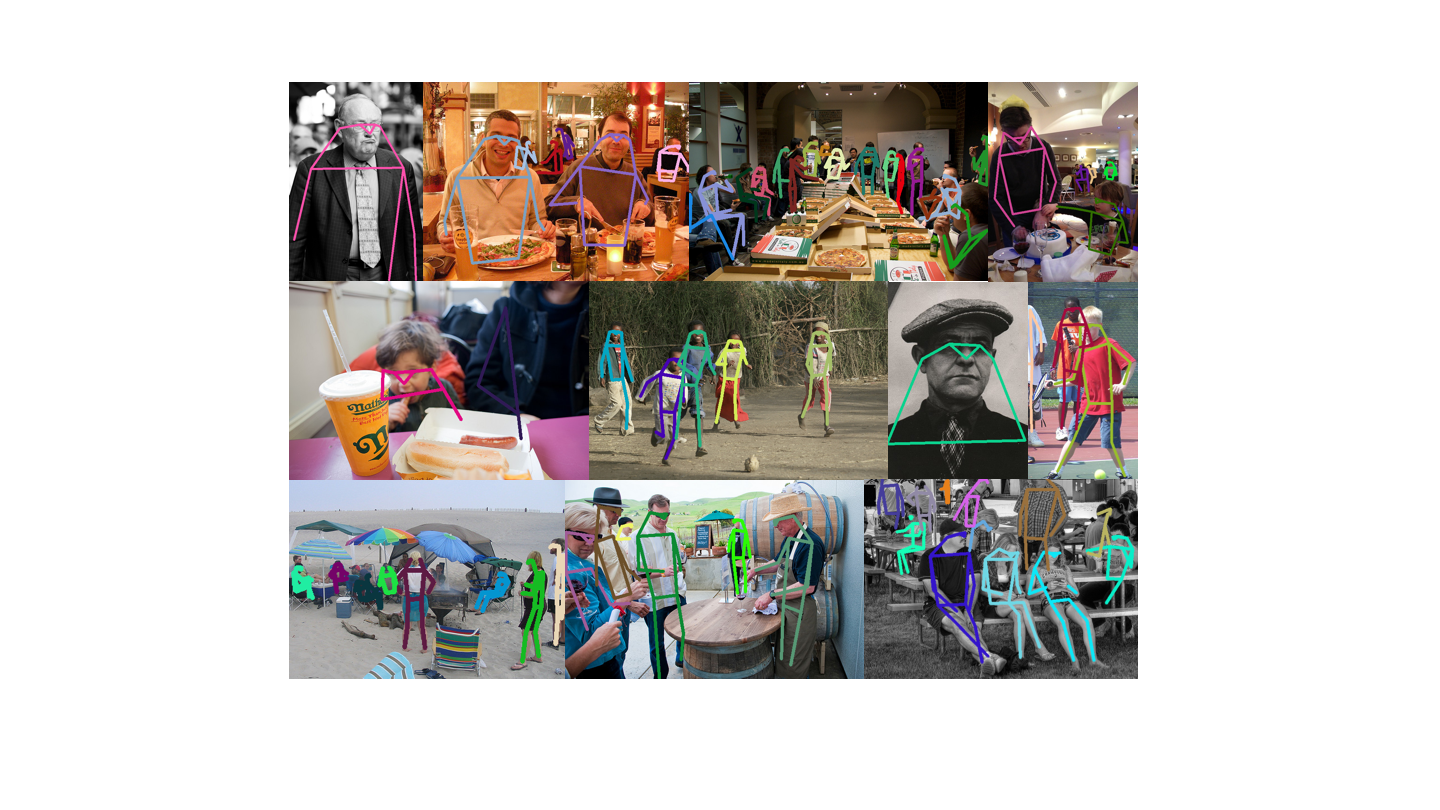}
\caption{Qualitative pose estimation results on MSCOCO validation images}
\label{fig:pose-exs}
\end{figure*}

\subsection{Instance Segmentation}

The goal of instance segmentation is to detect and classify object
instances while providing a segmentation mask for each object. As a
proof of concept we show how to apply our approach to this problem,
and demonstrate preliminary results. Like multi-person pose
estimation, instance segmentation is a problem of joint detection and
grouping. Pixels belonging to an object class are detected, and then
those associated with a single object are grouped together. For
simplicity the following description of our approach assumes only one
object category.

Given an input image, we use a stacked hourglass network to produce
two heatmaps, one for detection and one for tagging. The detection
heatmap gives a detection score at each pixel indicating whether the
pixel belongs to \emph{any} instance of the object category, that is,
the detection heatmap segments the foreground from background. At the
same time, the tagging heatmap tags each pixel such that pixels
belonging to the same object instance have similar tags.

To train the network, we supervise the detection heatmap by comparing
the predicted heatmap with the ground truth heatmap (the union of all
instance masks). The loss is the mean squared error between the two
heatmaps. We supervise the tagging heatmap by imposing a loss that
encourages the tags to be similar within an object instance and
different across instances. The formulation of the loss is similar to
that for multiperson pose. There is no need to do a comparison of
every pixel in an instance segmentation mask. Instead we randomly
sample a small set of pixels from each object instance and do pairwise
comparisons across the group of sampled pixels.

Formally, let $h \in R^{W \times H}$ be a predicted $W\times H$
tagging heatmap.  Let $x$ denote a pixel location and $h(x)$ the tag
at the location, and let $S_n={x_{kn}}, k=1,\ldots,K$ be a set of
locations randomly sampled within the $n$-th object instance.  The
grouping loss $L_g$ is defined as
\begin{multline*}
  L_g(h,T) = \sum_{n}\sum_{x\in S_n} \sum_{x'\in S_n} (h(x) - h(x'))^2 \\
  + \sum_{n}\sum_{n'} \sum_{x\in S_n} \sum_{x'\in S_{n'}}
  \exp\{-\frac{1}{2\sigma^2} (h(x) - h(x'))^2\}
\end{multline*}

To decode the output of the network, we first threshold on the
detection channel heatmap to produce a binary mask. Then, we look at
the distribution of tags within this mask. We calculate a histogram of
the tags and perform non-maximum suppression to determine a set of
values to use as identifiers for each object instance. Each pixel from
the detection mask is then assigned to the object with the closest tag
value. See Figure \ref{fig:inst-hm} for an illustration of this
process.

Note that it is straightforward to generalize from one object
category to multiple: we simply output a detection heatmap and a
tagging heatmap for each object category. As with multi-person pose,
the issue of scale invariance is worth consideration. Rather than
train a network to recognize the appearance of an object instance at
every possible scale, we evaluate at multiple scales and combine
predictions in a similar manner to that done for pose estimation.

\begin{figure*}
\centering
\includegraphics[width=\textwidth, clip]{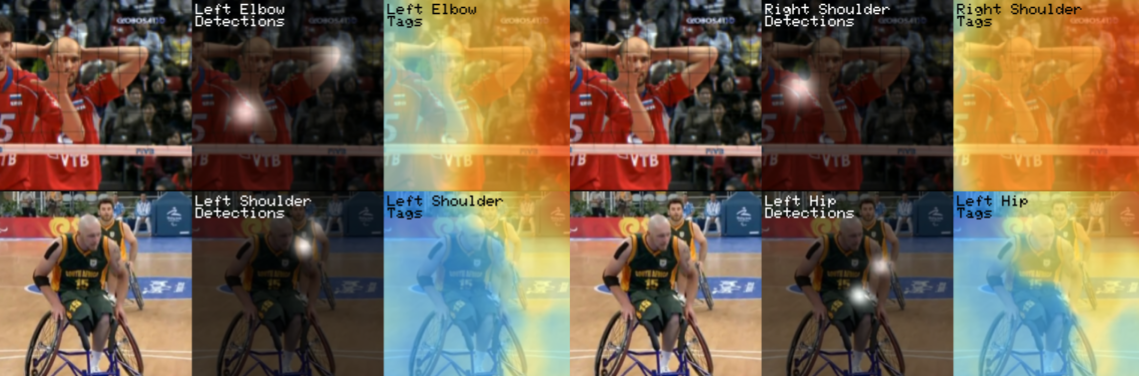}
\caption{Here we visualize the associative embedding channels for
  different joints. The change in embedding predictions across joints
  is particularly apparent in these examples where there is
  significant overlap of the two target figures.}
\label{fig:hmvis-2}
\end{figure*}

\begin{table*}
  \centering
  \begin{tabular}[t]{l||*{7}{c}|r} \hline
    &Head & Shoulder & Elbow & Wrist & Hip & Knee  & Ankle & Total\\ \hline
    %Varadarajan et al., '17~\cite{}& \bf{92.1} & 85.9 & 72.9 & 61.6 & 72.0 & 64.6 & 56.6 & 72.2
    Iqbal\&Gall, ECCV16~\cite{iqbal2016multi} & 58.4 & 53.9 & 44.5 & 35.0 & 42.2 &
    36.7 & 31.1 & 43.1 \\
    Insafutdinov et al., ECCV16~\cite{insafutdinov16ariv}& 78.4 & 72.5 & 60.2 & 51.0 & 57.2
    & 52.0 & 45.4 & 59.5 \\
    Insafutdinov et al., arXiv16a~\cite{pishchulin16cvpr}& 89.4 & 84.5 & 70.4 & 59.3 & 68.9
    & 62.7 & 54.6 & 70.0 \\ Levinkov et al.,
    CVPR17~\cite{levinkov2017joint}& 89.8 & 85.2 & 71.8 & 59.6 & 71.1
    & 63.0 & 53.5 & 70.6 \\ Insafutdinov et al.,
    CVPR17~\cite{insafutdinov2016articulated}& 88.8 & 87.0 & 75.9 &
    64.9 & 74.2 & 68.8 & 60.5 & 74.3 \\ Cao et al.,
    CVPR17~\cite{cao2016realtime}& 91.2 & 87.6 & 77.7 & 66.8 & 75.4 &
    68.9 & 61.7 & 75.6 \\ Fang et al., arXiv17~\cite{fang2016rmpe} &
    88.4 & 86.5 & 78.6 & \bf{70.4} & 74.4 & \bf{73.0} & \bf{65.8} &
    76.7 \\ \hline Our method & \bf{92.1} & \bf{89.3} & \bf{78.9} & 69.8 &
    \bf{76.2} & 71.6 & 64.7 & \textbf{77.5}
    \label{table:mpii}
    \\ \hline

  \end{tabular}

\caption{Results (AP) on MPII Multi-Person.}
\end{table*}

\section{Experiments}

\subsection{Multiperson Pose Estimation}
\smallskip
\noindent {\bf Dataset}
We evaluate on two datasets: MS-COCO~\cite{lin2014microsoft} and MPII Human
Pose~\cite{andriluka20142d}. MPII Human Pose consists of about 25k images and contains
around 40k total annotated people (three-quarters of which are
available for training). Evaluation is performed on MPII Multi-Person,
a set of 1758 groups of multiple people taken from the test set as
outlined in \cite{pishchulin16cvpr}. The groups for MPII
Multi-Person are usually a subset of the total people in a particular
image, so some information is provided to make sure predictions are
made on the correct targets. This includes a general bounding box and
scale term used to indicate the occupied region. No information is
provided on the number of people or the scales of individual
figures. We use the evaluation metric outlined by Pishchulin et
al. \cite{pishchulin16cvpr} calculating average precision of
joint detections.

\begin{table*}[t]
  \centering
  %\small
  \begin{tabular}[htp]{l||*{10}{c}}  \hline
              & AP & AP$^{50}$ & AP$^{75}$ & AP$^M$ & AP$^L$ & AR &
    AR$^{50}$ & AR$^{75}$ & AR$^M$ & AR$^L$\\ \hline
    CMU-Pose \cite{cao2016realtime} & 0.611 & 0.844 & 0.667 & 0.558 & 0.684 &
    0.665 & 0.872 & 0.718 & 0.602 & 0.749\\
    G-RMI \cite{papandreou2017towards} & 0.643 & 0.846 & 0.704 & \bf{0.614}
    & 0.696 & 0.698 & 0.885 & 0.755 & 0.644 & 0.771\\ \hline
    Our method & \bf{0.663} & \bf{0.865} & \bf{0.727} & 0.613 & \bf{0.732}
    & \bf{0.715} & \bf{0.897} & \bf{0.772} & \bf{0.662} & \textbf{0.787} \\ \hline
  \end{tabular}
  \caption{Results on MS-COCO \textbf{test-std}, excluding systems trained with external
    data.}
  \label{table:test-std}
\end{table*}

\begin{table*}[t]
  \centering
  %\small
  \begin{tabular}[t]{l||*{10}{c}} \hline
    & AP & AP$^{50}$ & AP$^{75}$ & AP$^M$ & AP$^L$ & AR & AR$^{50}$ &
    AR$^{75}$ & AR$^M$ & AR$^L$\\ \hline
%              & AP & AP.5 & AP.75 & AP(M) & AP(L) & AR & AR.5 & AR.75
%              & AR(M) & AR(L)\\ \hline
    CMU-Pose \cite{cao2016realtime} & 0.618 & 0.849 & 0.675 & 0.571 &
    0.682 & 0.665 & 0.872 & 0.718 & 0.606 & 0.746\\
    %RMPE & 0.618 & 0.837 & 0.698 & 0.586 & 0.673 & 0.676 & 0.875 &
    %0.746 & 0.630 & 0.740\\
    Mask-RCNN \cite{he2017mask} & 0.627 & \bf{0.870} & 0.684 & 0.574 & 0.711 & -- & -- & -- & -- & -- \\
    G-RMI \cite{papandreou2017towards} & 0.649 & 0.855 & 0.713 & \bf{0.623} & 0.700 & 0.697 & 0.887 & 0.755 & 0.644 & 0.771 \\ \hline
    Our method & \bf{0.655} & 0.868 & \bf{0.723} & 0.606 & \bf{0.726} & \bf{0.702} & \bf{0.895} & \bf{0.760} & \bf{0.646} & \textbf{0.781} \\ \hline
  \end{tabular}
  \caption{Results on MS-COCO \textbf{test-dev}, excluding systems
    trained with external data.}
  \label{table:test-dev}
\end{table*}

MS-COCO~\cite{lin2014microsoft} consists of around 60K training images
with more than 100K people with annotated keypoints. We report
performance on two test sets, a development test set (test-dev) and a
standard test set (test-std). We use the official evaluation metric
that reports average precision (AP) and average recall (AR) in a
manner similar to object detection except that a score based on
keypoint distance is used instead of bounding box overlap. We refer
the reader to the MS-COCO website for details~\cite{cocowebsite}.

\smallskip
\noindent {\bf Implementation} The network used for this task consists
of four stacked hourglass modules, with an input size of $512\times
512$ and an output resolution of $128\times128$. We train the network
using a batch size of 32 with a learning rate of 2e-4 (dropped to 1e-5
after 100k iterations) using
Tensorflow~\cite{tensorflow2015-whitepaper}. The associative embedding
loss is weighted by a factor of 1e-3 relative to the MSE loss of the
detection heatmaps. The loss is masked to ignore crowds with sparse
annotations. At test time an input image is run at multiple scales;
the output detection heatmaps are averaged across scales, and the tags
across scales are concatenated into higher dimensional tags. Since the
metrics of MPII and MS-COCO are both sensitive to the precise
localization of keypoints, following prior
work~\cite{cao2016realtime}, we apply a single-person pose
model~\cite{newell2016stacked} trained on the same dataset to further
refine predictions.

\smallskip
\textbf{MPII Results}
%{\bf MPII Results} 
Average precision results can be seen in Table 1 %\ref{table:mpii}
demonstrating an improvement over state-of-the-art methods in overall
AP. Associative embedding proves to be an effective method for
teaching the network to group keypoint detections into individual
people. It requires no assumptions about the number of people present
in the image, and also offers a mechanism for the network to express
confusion of joint assignments. For example, if the same joint of two
people overlaps at the exact same pixel location, the predicted
associative embedding will be a tag somewhere between the respective
tags of each person.

We can get a better sense of the associative embedding output with 
 visualizations of the embedding heatmap (Figure
\ref{fig:hmvis-2}). We put particular focus on the difference in the
predicted embeddings when people overlap heavily as the severe
occlusion and close spacing of detected joints make it much more
difficult to parse out the poses of individual people.

\textbf{MS-COCO Results} Table~\ref{table:test-std} and
Table~\ref{table:test-dev} report our results on MS-COCO. We report
results on both test-std and test-dev because not all recent methods
report on test-std. We see that on both sets we achieve the state of
the art performance. An illustration of the network's predictions can
be seen in Figure \ref{fig:pose-exs}. Typical failure cases of the network stem from overlapping
  and occluded joints in cluttered scenes. Table \ref{table:test-dev-self} reports 
performance of ablated versions of our full pipeline, showing the
contributions from applying our model at multiple scales and from
further refinement using a single-person pose estimator. We see that
simply applying our network at multiple scales already achieves
competitive performance against prior state of the art methods,
demonstrating the effectiveness of our end-to-end joint detection and
grouping.

We also perform an additional experiment on MS-COCO to gauge the relative difficulty of
detection versus grouping, that is, which part is the main bottleneck of our system. We evaluate
our system on a held-out set of 500 training images. In this evaluation, 
we replace the predicted detections with the ground truth detections but still use
the predicted tags. Using the ground truth detections improves AP from 59.2 to 94.0. This
shows that keypoint detection is the main bottleneck of our system, whereas the network
has learned to produce high quality grouping. This fact is also supported by qualitative inspection of
the predicted tag values, as shown in Figure~\ref{fig:ae-vs-kpt}, from which we can see that
the tags are well separated and decoding the grouping is straightforward. 

\newcolumntype{L}[1]{>{\hsize=#1\hsize\raggedright\arraybackslash}X}%
\newcolumntype{R}[1]{>{\hsize=#1\hsize\raggedleft\arraybackslash}X}%
\newcolumntype{C}[1]{>{\hsize=#1\hsize\centering\arraybackslash}X}%

\begin{table}
  \centering
  \small
  %\begin{tabularx}[t]{l||ccccc}
  \begin{tabularx}{\textwidth}{L{2}|| C{.5} C{.5} C{.5} C{.5} C{.5}}
    \hline
            & AP    & AP$^{50}$ & AP$^{75}$ & AP$^M$ & AP$^L$ \\ \hline
    single scale & 0.566 & 0.818 & 0.618  & 0.498 & 0.670\\ \hline
    single scale + refine & 0.628 & 0.846 & 0.692  & 0.575 & 0.706\\ \hline
    multi scale           & 0.630 & 0.857 & 0.689  & 0.580 & 0.704\\ \hline
    multi scale  + refine & 0.655 & 0.868 & 0.723  & 0.606 & 0.726 \\ \hline
  \end{tabularx}
   \caption{Effect of multi-scale evaluation and
    single person refinement on MS-COCO
    \textbf{test-dev}.}
  \label{table:test-dev-self}
\end{table}

\subsection{Instance Segmentation}

\smallskip
\noindent
{\bf Dataset} For evaluation we use the $val$ split of PASCAL VOC 2012
\cite{everingham2015pascal} consisting of 1,449 images.
Additional pretraining is done with images from MS COCO
\cite{lin2014microsoft}. Evaluation is done using mean average
precision of instance segments at different IOU thresholds.
\cite{hariharan2014simultaneous,dai2016instance,liu2016mpa}

\smallskip
\noindent
{\bf Implementation} The network is trained in Torch \cite{torch7}
with an input resolution of $256 \times 256$ and output resolution of
$64 \times 64$. The weighting of the associative embedding loss is
lowered to 1e-4. During training, to account for scale, only objects
that appear within a certain size range ar supervised, and a loss mask
is used to ignore objects that are too big or too small. In PASCAL
VOC ignore regions are also defined at object boundaries, and we
include these in the loss mask. Training is done from scratch on MS
COCO for three days, and then fine tuned on PASCAL VOC $train$ for 12
hours. At test time the image is evaluated at 3-scales (x0.5, x1.0,
and x1.5). Rather than average heatmaps we generate instance proposals
at each scale and do non-maximum suppression to remove overlapping
proposals across scales. A more sophisticated approach for multi-scale
evaluation is worth further exploration.

\begin{figure}
  \ffigbox[\textwidth]{\includegraphics[width=\textwidth, trim={0cm .2cm 0cm .2cm}, clip]{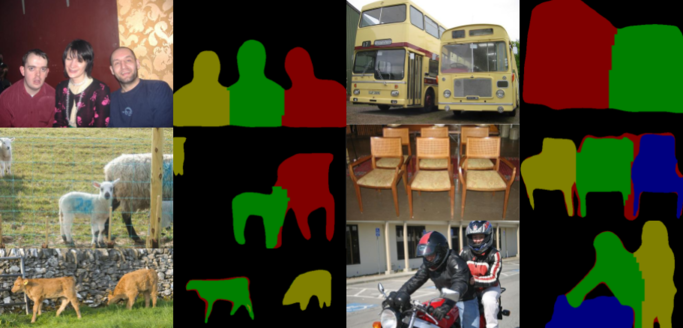}
      \caption{Example instance predictions produced by our system on the PASCAL
VOC 2012 validation set.}
  \label{fig:inst-exs}
}

\end{figure}

\smallskip
\noindent
{\bf Results} We show mAP results on the $val$ set of PASCAL VOC 2012
in Table \ref{table:inst} along with some qualitative examples in
Figure \ref{fig:inst-exs}. We offer these results as a proof of
concept that associative embeddings can be used in this manner. We
achieve reasonable instance segmentation predictions using the
supervision as we use for multi-person pose. Tuning of training and
postprocessing will likely improve performance, but the main takeaway
is that associative embedding serves well as a general technique for
disparate computer vision tasks that fall under the umbrella of
detection and grouping problems.

\begin{table}
%\capbtabbox{%
  \begin{tabular}[t]{l||*{1}{c}|c}
    \hline
    & IOU=0.5 & IOU=0.7 \\ \hline
    SDS \cite{hariharan2014simultaneous} & 49.7 & 25.3 \\
    Hypercolumn \cite{hariharan2015hypercolumns} & 60.0 & 40.4 \\
    CFM \cite{dai2015convolutional} & 60.7 & 39.6 \\
    MPA \cite{liu2016mpa} & 61.8 & -- \\
    MNC \cite{dai2015instance} & 63.5 & 41.5 \\ \hline
    Our method & 35.1 & 26.0
    \label{table:inst}
    \caption{Semantic instance segmentation results (mAP) on PASCAL VOC 2012
      validation images.}
  \end{tabular}
\end{table}

\section{Conclusion}

In this work we introduce associative embeddings to supervise a
convolutional neural network such that it can simultaneously generate
and group detections. We apply this method to two vision problems:
multi-person pose and instance segmentation. We demonstrate the
feasibility of training for both tasks, and for pose we achieve
state-of-the-art performance. Our method is general enough to be
applied to other vision problems as well, for example multi-object
tracking in video. The associative embedding loss can be implemented
given any network that produces pixelwise predictions, so it can be
easily integrated with other state-of-the-art architectures.

{\small
\bibliographystyle{plain}
\bibliography{egbib}
}

\end{document}